\documentclass[conference]{IEEEtran}
\IEEEoverridecommandlockouts
\usepackage{cite}
\usepackage{amsmath,amssymb,amsfonts}
\usepackage{algorithmic}
\usepackage{graphicx}
\usepackage{textcomp}
\usepackage{xcolor}
\usepackage{stfloats}
\usepackage{pifont}
\usepackage{array}
\usepackage[utf8]{inputenc}
\usepackage{booktabs}
\usepackage{array}
\usepackage{soul}
\usepackage{xcolor}
\usepackage{float}
\usepackage{soul}
\usepackage{url}

\definecolor{pastelyellow}{RGB}{253,253,150}
\definecolor{pastelgreen}{RGB}{150,253,150}
\definecolor{pastelred}{RGB}{253,150,150}

\sethlcolor{pastelyellow} 

\def\BibTeX{{\rm B\kern-.05em{\sc i\kern-.025em b}\kern-.08em
    T\kern-.1667em\lower.7ex\hbox{E}\kern-.125emX}}
\begin{document}

\title{LLM-Guided Indoor Navigation with Multimodal Map Understanding
\thanks{This work was founded by European Union - Next Generation EU, in the context of The National Recovery and Resilience Plan, Investment Partenariato Esteso PE8 "Conseguenze e sfide dell'invecchiamento", Project Age-IT, CUP: B83C22004880006}
}

\author{\IEEEauthorblockN{
Alberto Coffrini\IEEEauthorrefmark{1}\IEEEauthorrefmark{2}, Paolo Barsocchi\IEEEauthorrefmark{1}, Francesco Furfari\IEEEauthorrefmark{1}, Antonino Crivello\IEEEauthorrefmark{1}, Alessio Ferrari\IEEEauthorrefmark{1}\IEEEauthorrefmark{3}}

\IEEEauthorblockA{
\IEEEauthorrefmark{1}\emph{Institute of Information Science and Technologies (ISTI)}, \emph{National Research Council of Italy (CNR)}, Pisa, Italy \\
\IEEEauthorrefmark{2}\emph{Department of Computer Science}, \emph{University of Pisa}, Pisa, Italy \\
\IEEEauthorrefmark{3}\emph{University College Dublin (UCD), School of Computer Science}, Belfield, Dublin 4, Ireland\\
}}

\maketitle

\begin{abstract}
Indoor navigation presents unique challenges due to complex layouts and the unavailability of GNSS signals. Existing solutions often struggle with contextual adaptation, and typically require dedicated hardware. In this work, we explore the potential of a Large Language Model (LLM), i.e., ChatGPT, to generate natural, context-aware navigation instructions from indoor map images. We design and evaluate test cases across different real-world environments, analyzing the effectiveness of LLMs in interpreting spatial layouts, handling user constraints, and planning efficient routes. Our findings demonstrate the potential of LLMs for supporting personalized indoor navigation, with an average of 86.59\% correct indications and a maximum of 97.14\%. The proposed system achieves high accuracy and reasoning performance. These results have key implications for AI-driven navigation and assistive technologies.

\end{abstract}

\begin{IEEEkeywords}
Large Language Model, Multimodal, AI-driven path planning, Indoor Navigation, Accessibility, Spatial Reasoning, Indoor Maps.
\end{IEEEkeywords}

\section{Introduction}
Efficient indoor navigation and scene representation are essential for various applications, including robotics, smart environments, and assistive technologies. Unlike outdoor navigation, which benefits from well-established global positioning systems (GPS) and extensive mapping databases, indoor navigation presents unique challenges \cite{torres2021towards}.  The complexity and variability of enclosed spaces, the absence of GPS signals, and the need for detailed real-time scene understanding make indoor navigation a demanding task \cite{furfari2019next}.
 
Current solutions rely on advanced computational models to interpret the environment and adapt to dynamic conditions \cite{shao2024moc}. However, they often lack flexibility, generalization capability, and require the support of dedicated hardware. 

There is a growing need for a system capable of providing personalized natural language navigation assistance to address these limitations\cite{buzzi2024chatbot}. Such a system should interpret indoor maps and generate human-like, context-aware navigation instructions tailored to individual users. 

%
People visiting a new location, such as airports, shopping malls, hospitals, or office buildings, often struggle with orientation, requiring clear and accessible guidance to reach their destinations efficiently. A common approach to overcoming these challenges is to ask other people for directions, who typically provide landmark-based instructions (e.g., \textit{you will see a shoe shop, turn right, then you will encounter a phone charging area...}). 

Given these challenges, leveraging Large Language Models (LLMs) for indoor navigation presents a promising alternative. LLMs can interpret complex multimodal inputs, reason over spatial relationships, and generate human-like natural language indications tailored to user-specific queries. This adaptability makes them well-suited for generating dynamic, human-like navigation instructions based on indoor maps\cite{tsai2023multimodal}.

In fact, recent advances in LLMs have opened new avenues for user-friendly navigation solutions for indoor and outdoor contexts, and preliminary studies have been performed leveraging LLMs to address this goal.  

For example, in \cite{fang2024travellmplannewpublic}, the authors introduce \emph{TraveLLM}, a system for public transit route planning. Their study compares multiple LLMs---including ChatGPT-4, Claude 3, and Gemini---demonstrating that these models can integrate real-time data and user-specific constraints to form contextually aware navigation recommendations than conventional platforms (e.g., Google Maps). 
However, their study focuses on outdoor scenarios, leaving \textit{indoor} navigation unexplored.

Meanwhile, Zhang et al. \cite{10.1145/3688828.3699636} focus on improving navigation for people with visual impairments (PVI). \emph{NaviGPT} integrates LiDAR-based obstacle detection, vibration feedback, and LLM-generated instructions to produce a real-time user experience. Unlike existing tools, e.g., Be My AI\footnote{\url{https://www.bemyeyes.com/blog/introducing-be-my-ai}} and Seeing AI\footnote{\url{https://www.seeingai.com/}}, which require multiple apps for obstacle detection, image recognition, or contextual directions, NaviGPT consolidates these capabilities into one cohesive system. This seamless integration offers continuous feedback without user distraction from frequent application switching. However, no evaluation of the system has been provided, and at the current stage, the proposal appears to mainly target outdoor environments.

For indoor context, in \cite{app14209343}, the authors examine the development of \emph{geo-descriptions}---textual narratives that describe the internal layout of a building. These authors combine classical GIS spatial analysis and LLM-based text generation to automate the creation of human-readable indoor maps, which is traditionally labour-intensive. 
However, the authors do not consider querying the system for navigation paths in the reconstructed textual representation of the building. Furthermore, the evaluation is performed on simplified cases.

Despite these preliminary studies, the use of visual maps as input for LLMs to provide indications based on user requests has not been explored. This could highly simplify the architecture of indoor navigation systems, thus making them easily adaptable to different indoor environments. Once a visual map is made available---as is common for airports, shopping malls, and large environments with several points of interest---a system based on these principles is expected to be ready to use in any environment without complex, \textit{ad hoc}, configurations. Indeed, this approach avoids the task of providing structured representations of the maps for each environment, which can be time-consuming. 

To address the research gap, we evaluate the potential of an LLM, namely ChatGPT-4 (in two versions, ChatGPT-4o and the superior o3), to assist in indoor navigation using three real-world interior 2D maps of buildings, including two airport terminal maps and one shopping mall map. 

We investigate the LLM ability to generate step-by-step guidance based on diverse user preferences and spatial constraints. 

Using different real scenarios, we want to evaluate different test cases across three real-world environments, analyzing the effectiveness of LLMs in interpreting spatial layouts, handling user constraints, and planning efficient routes.

The main contributions of this paper are as follows:
\begin{enumerate}
    \item We evaluate the effectiveness of ChatGPT-4 in generating accurate, human-like navigation guidance based on visual indoor maps. 
    \item We design test cases to assess LLM performance under different conditions, such as long distances, extensive textual annotations, and complex map layouts.
    \item According to \cite{anagnostopoulos2025ordip}, all supplementary materials, including prompts, input maps, annotation guide, and results, are publicly available\footnote{\url{https://doi.org/10.5281/zenodo.14993277}}.
\end{enumerate}

This work explores the role of LLMs in indoor navigation, aiming to bridge the gap between traditional map-based systems and intelligent, real-time, and user-adaptive guidance solutions.




\begin{figure}[t]
  \centering
  \includegraphics[width=0.5\textwidth]{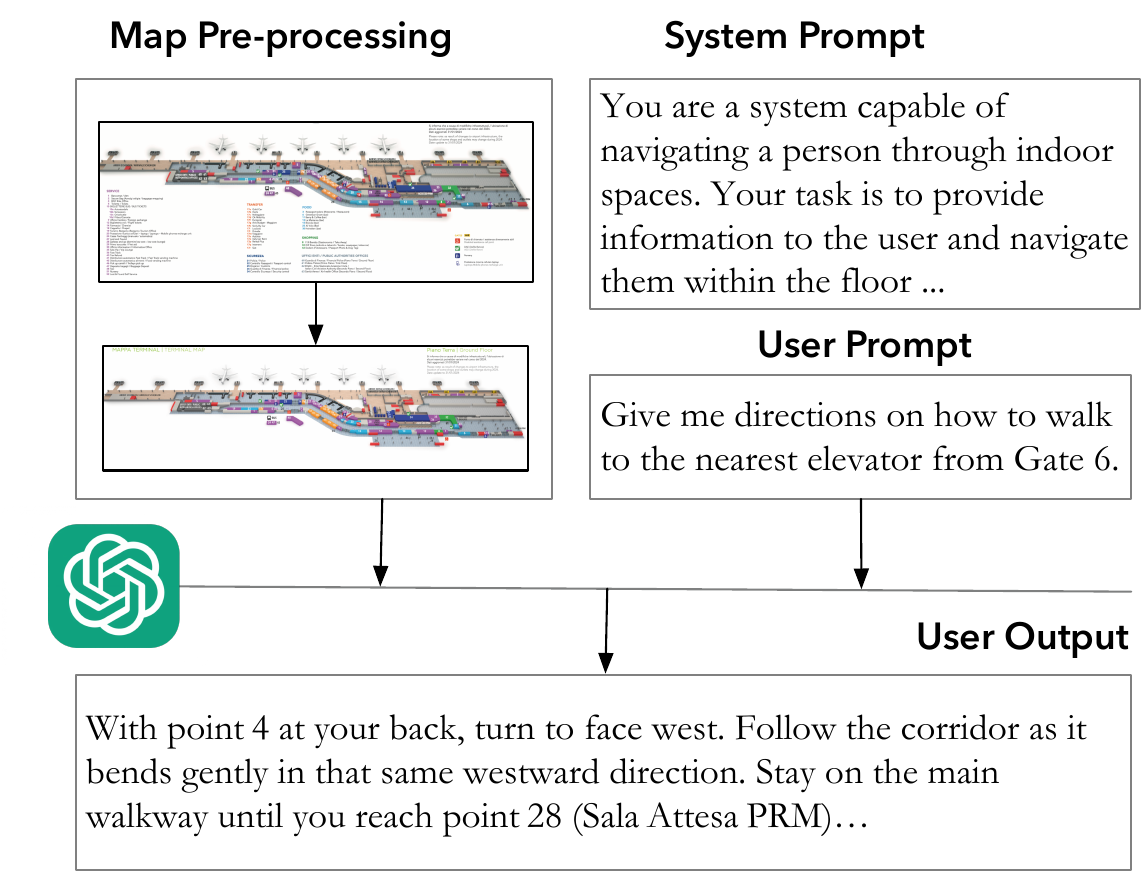}
  \caption{Flowgraph of the experimental setting.}
  \label{fig:flowgraph}
\end{figure}




\section{System Overview and Pipeline Description} 
\label{pipeline}
This section illustrates the approach we followed to evaluate the capability of the LLM to support indoor navigation. Given a user query, we expect the LLM to deliver a detailed output, including step-by-step directions, points of interest (POIs), and potential obstacles, all aimed at enhancing the user's navigation experience.

We performed three different experiments, considering five different types of queries, and three maps. Each experiment corresponds to an incremental refinement, e.g., different pre-processing steps or model versions, introduced to overcome the limitations of the previous experiment. All the experiments are based on the same pipeline as described in the following.

Preliminary tests were conducted using several state-of-the-art LLM frontier models, including Gemini, ChatGPT, and Claude. Based on these initial evaluations, the model that delivered the most promising results—ChatGPT—was selected for the experiments presented in this study.

Figure~\ref{fig:flowgraph} describes the pipeline of our experimental process. Its four main components are:
\begin{itemize}
    \item \textit{Map Pre-processing} The input images were manually adjusted to enhance performance. A range of Pre-processing techniques (e.g., removal of the legend, simplification of the map) was applied throughout the different experiments, each informed by specific hypotheses or objectives.
    Detailed descriptions of the manual pre-processing steps are provided in each experiment analysis section, i.e., Section \ref{sec:test1}, \ref{sec:test2}, and \ref{sec:test3}.

    \item \textit{System Prompt} The system prompt is hard-coded for the language model and is applied uniformly to all user queries. It contains comprehensive instructions, relevant details, and guidelines for avoiding potential errors. This prompt is the result of several iterative refinements aimed at preventing critical and systematic mistakes, and enhancing the model's overall performance. Across the various experiments, two different versions of the system prompt were employed. The specific version used in each experiment is disclosed and discussed in the corresponding experiment analysis sections.
    \item \textit{User Prompt} The user makes a request, i.e., an LLM user prompt, aiming to navigate from point A to point B on the map. 
    \item \textit{User Output} The system produces an output with human-like indications delivered in written form. 
\end{itemize}

The model receives comprehensive contextual information by integrating the user prompt, system prompt, and sanitized map image, enabling it to generate navigational instructions. This approach ensures that the system is not only responsive to user inputs but also adaptive to various indoor environments, making it versatile across different settings.

A single user prompt structure---with different content---was used across all images in an experiment. The structure does not vary between maps and does not include detailed descriptions specific to any map. Instead, references to locations in the user prompts are made either by the numerical ID of a point of interest or by the point's name, depending on the map's representation of the points. It should be highlighted that, in a practical scenario with a real user querying the LLM, prompt engineering is not feasible. 

\section{Experimental Set-up}
\label{sec:experiment}

The experiments were conducted using three different maps, all in high quality .png format:

\begin{itemize}
    \item \textit{Bergamo} The Terminal map is visually clear and well-organized, with labeled POIs and a straightforward layout, but the high density and small size of elements hinder accurate image parsing by the language model.
    
    \item \textit{Bologna} Similar to Bergamo airport, this map has a central corridor flanked by shops, but it differs in its color scheme, floor representation, and a more curved, less linear layout.
    
    \item \textit{Orio Center} This shopping mall map was selected for its complex layout, featuring frequent directional changes and intersecting corridors, with small retail outlets embedded in the paths, making spatial interpretation more challenging.
\end{itemize}

For each map, five requests are made, each with a different category of difficulty. These were defined \textit{ad hoc} based on a visual analysis of each map. The first three queries differ in terms of distance between starting and destination points, i.e., short (S), medium (M), and long (L). The other two queries, referred to as ``specific requests'' (Spec), explore other elements of complexity, which depend on each map. Specifically, we challenged the LLM with routes that have hidden or poorly marked IDs of POIs, or with prompts in which the specific destination was not explicitly stated, e.g., \textit{``Show me how to walk to the nearest dining area from gate number 26"}. In this case, the query requires a more detailed search near gate 26 to locate the closest dining area. Table~\ref{tablequery} reports examples of user prompts for the Bologna Airport. 

Each reference to query number \textit{X} of a specific map will be cited as follows: \textit{BER-X} for Bergamo airport, \textit{BOL-X} for Bologna airport, and \textit{ORIO-X} for Orio Center.

\begin{table*}[!ht]
\centering
\caption{User prompt examples for the Bologna Airport.}
\begin{tabular}{|p{1.5cm}|p{2cm}|p{6cm}|p{6cm}|}
\hline
\textbf{Scenario ID} & \textbf{Query Category} & \textbf{User Prompt} & \textbf{Query Rationale} \\
\hline
BOL-1 & Short & Give me directions on how to walk to the Vecchia Bologna Osteria (4 on the map) from the Carisbo (11 on the map). & Testing on a short path \\
\hline
BOL-2 & Medium & Give me directions on how to walk to the Carrefour (15 on the map) from the Mondadori (7 on the map).
 & Testing on a medium path. \\
\hline
BOL-3 & Long & Give me directions on how to walk to the point 2 on the map from the point 3 on the map. & Testing on a long path. 
\\
\hline
BOL-4 & Spec & Give me directions on how to walk to the closest toilet on the map from point 3.  & Testing the ability to find the closest toilet and recognize the related icon.\\
\hline
BOL-5 & Spec & Give me directions on how to walk to the closest taxi station from the First Aid on the map & Testing the ability to recognize difficult icons on the map and find the closest path.\\
\hline
\end{tabular}
\label{tablequery}
\end{table*}

\subsection{Evaluation Method}
The evaluation of the LLM’s responses is carried out by defining a ground truth of correct paths. Once the start and destination points are established for each query, we manually identify the optimal route by imagining a user standing at the starting point and then considering the shortest, least convoluted path through the structure.

Each instruction provided by the model---i.e., the description of a path segment---is assigned a label reflecting its accuracy, which the authors decided by comparing the ground truth and the LLM output. There are two labels:

\begin{itemize}
    \item \textit{Correct} – the instruction reflects a segment of the correct route(s) of the ground truth.  
    \item \textit{Incorrect} – the instruction does not reflect the actual state of the navigation or provides misleading information to the user, such as indicating the wrong direction or suggesting physically impossible paths (e.g., through walls). 
\end{itemize}

\begin{figure}[!ht]
  \centering
  \includegraphics[width=0.5\textwidth]{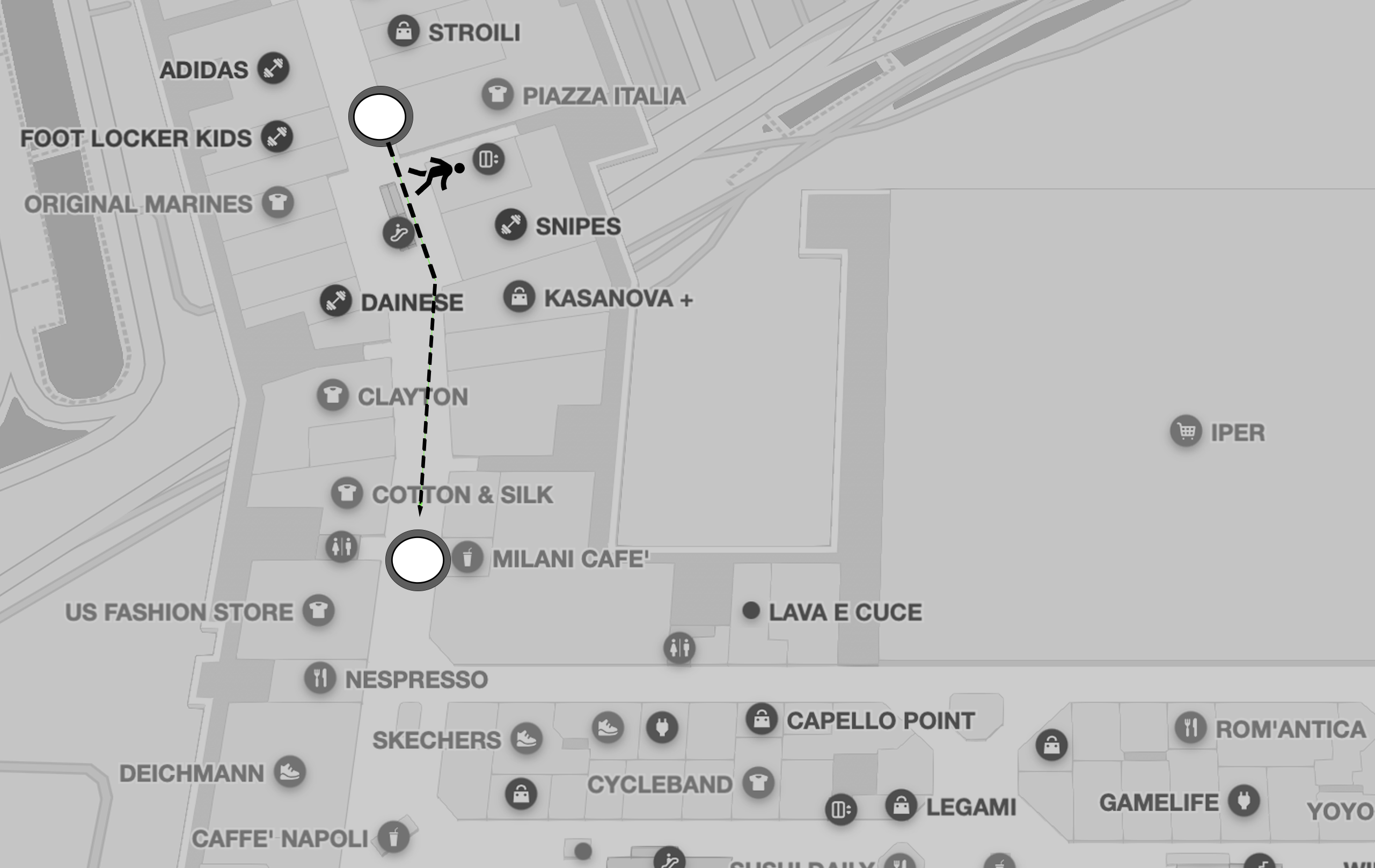}
  \caption{Correct path considered for ORIO-2 query (Table~\ref{tablequery}) on the Orio Center mall map. }
  \label{fig:orio_path}
\end{figure}

In the supplementary material, we share the annotation guide adopted to identify path segments and label them, to ensure the replicability of our experiments. In the following, we consider one example to showcase correct and incorrect cases and the rationale for our tagging. 
Taking the query ``How to go from Piazza Italia to Milani Cafè''---cf. Figure~\ref{fig:orio_path} for the area of interest and the correct path---this yields the following output:

``To get to Milani Cafè from Piazza Italia, follow these steps:
    \begin{itemize}
        \item  From Piazza Italia \sethlcolor{pastelgreen} \hl{head south} \sethlcolor{pastelgreen} \hl{ towards Snipes.}
        \item \sethlcolor{pastelgreen} \hl{Continue straight past Snipes,} \sethlcolor{pastelred} \hl{and you will see Clayton} \sethlcolor{pastelred} \hl{ on your left.}
        \item \sethlcolor{pastelgreen} \hl{Keep walking straight and pass Dainese,} \sethlcolor{pastelred} \hl{which will also be on your left.}
        \item \sethlcolor{pastelgreen} \hl{Milani Cafè is just ahead} \sethlcolor{pastelred} \hl{on the right side,} \sethlcolor{pastelred} \hl{next to Cotton \& Silk.}''
    \end{itemize}

The green (correct) and red (incorrect) highlights are provided according to the following rationale: 
\begin{itemize}
\item Green directions like \sethlcolor{pastelgreen} \hl{``continue straight past Snipes''} are considered totally correct as they align with the correct path. The same is true for \sethlcolor{pastelgreen} \hl{``keep walking straight and pass Dainese''}, and the other instructions. 
\item Instead, instructions such as \sethlcolor{pastelred} \hl{``and you will see Clayton on your left''} and \sethlcolor{pastelred} \hl{``which will also be on your left''} are incorrect because Clayton is actually located on the right and after Dainese. Errors related to order and left/right orientation are particularly misleading, as they may cause the user to question whether the correct direction is reversed.
\end{itemize}

It should be noted that, in practice, one incorrect indication may jeopardize the entire path. For example, users may reverse their direction based on a single incorrect instruction and subsequently get lost, even if the following directions are correct. However, in this preliminary evaluation phase, we decided to focus our evaluation on individual instructions rather than the entire path to assess the accuracy of each step independently and better isolate specific types of errors.

\section{Experimental Results}
This section presents and analyzes three individual experiments in detail. For each experiment, the complete pipeline is described step by step. Summary tables provide a comprehensive overview of success and failure rates across different queries and map configurations. The section concludes with a detailed discussion of common errors encountered and the improvements made over the previous version of the pipeline.

\subsection{Experiment I - Original Image Map}
\label{sec:test1}
The first experiment was performed with ChatGPT 4o. The complete and detailed analysis of this experiment is described in \cite{coffrini2025methodllmenabledindoornavigation}.
\begin{itemize}
    \item \textit{Pre-processing}: Removed the legend from the image, to ensure that the input map is free of noise that can affect the LLM performance.
    \item \textit{System prompt}: defined as '\textit{SP\#1}', was designed based on the characteristics of the input image. Before data collection, several improvements were made to the prompt to enhance the model’s overall performance. After various tests, it became evident that the LLM's output exhibited recurring errors, e.g., evaluating the proximity between points based on the numerical difference between their IDs on the map. These prompt refinements were introduced specifically to address those issues and to better guide the model's reasoning in relation to the visual context.
\end{itemize}

In Table \ref{first_test_table}, a total of 186 output indications were evaluated, with 92 incorrect (49.64\%), and 94 correct (50.54\%). The highest score of correct indications is for Bologna Airport map (59.95\%) and the lowest result is for Bergamo Airport map (37.74\%).
These results suggest that clearer and less cluttered visual representations lead to better model performance. Notably, the Bologna map contains fewer facilities and additional information, which likely contributed to the improved navigation accuracy.

\begin{table*}[!ht]
\centering
\caption{Results with Image without legend and System prompt SP\#1 with ChatGPT-4o}
\label{first_test_table}
\begin{tabular}{lcccccc}
\toprule
\textbf{Query} 
& \multicolumn{2}{c}{\textbf{BERGAMO}} 
& \multicolumn{2}{c}{\textbf{BOLOGNA}} 
& \multicolumn{2}{c}{\textbf{ORIO}} \\
\cmidrule(lr){2-3} \cmidrule(lr){4-5} \cmidrule(lr){6-7}
& \textbf{Incorrect} & \textbf{Correct} 
& \textbf{Incorrect} & \textbf{Correct} 
& \textbf{Incorrect} & \textbf{Correct} \\
\midrule
Query-1 (S)     & 7 (87.50\%) & 1 (12.50\%) & 8 (47.06\%) & 9 (52.95\%) & 5 (71.43\%) & 2 (28.57\%)\\
Query-2 (M)     & 7 (70.00\%) & 3 (30.00\%) & 4 (28.57\%) & 10 (71.43\%) & 5 (50.00\%) & 5 (50.00\%)\\
Query-3 (L)     & 12 (54.55\%) & 10 (45.45\%) & 6 (42.86\%) & 8 (57.14\%) & 4 (22.22\%) & 14 (77.78\%) \\
Query-4 (Spec)  & 3 (60.00\%) & 2 (40.00\%) & 5 (62.50\%) & 3 (37.50\%) & 13 (76.47\%) & 4 (23.53\%) \\
Query-5 (Spec)  & 4 (50.00\%) & 4 (50.00\%) & 4 (30.77\%) & 9 (69.23\%) & 5 (33.33\%) & 10 (66.67\%)\\
\midrule
\textbf{Total}  
& 33 \textbf{(62.26\%)} & 20 \textbf{(37.74\%)} 
& 27 \textbf{(40.91\%)} & 39 \textbf{(59.09\%)} 
& 32 \textbf{(47.76\%)} & 35 \textbf{(52.24\%)}\\
\bottomrule
\end{tabular}
\end{table*}

Furthermore, we noticed that the complexity of the user query and the layout of the map do not appear to strongly influence the performance. Instead, these appear to be affected by the visual characteristics of the maps, e.g., the degree of contrast between colours and size and the information overload due to the density of POIs.

\subsection{Experiment II - Graph Style Map}
\label{sec:test2}
This experiment aims to enhance the performance of the LLM, particularly in identifying the correct path and accurately recognizing POIs and other facilities based on the user’s perspective. It specifically addresses the most frequent errors observed in the previous experiments.

\begin{figure}[t]
  \centering
  \includegraphics[width=0.5\textwidth,height=7cm,keepaspectratio]{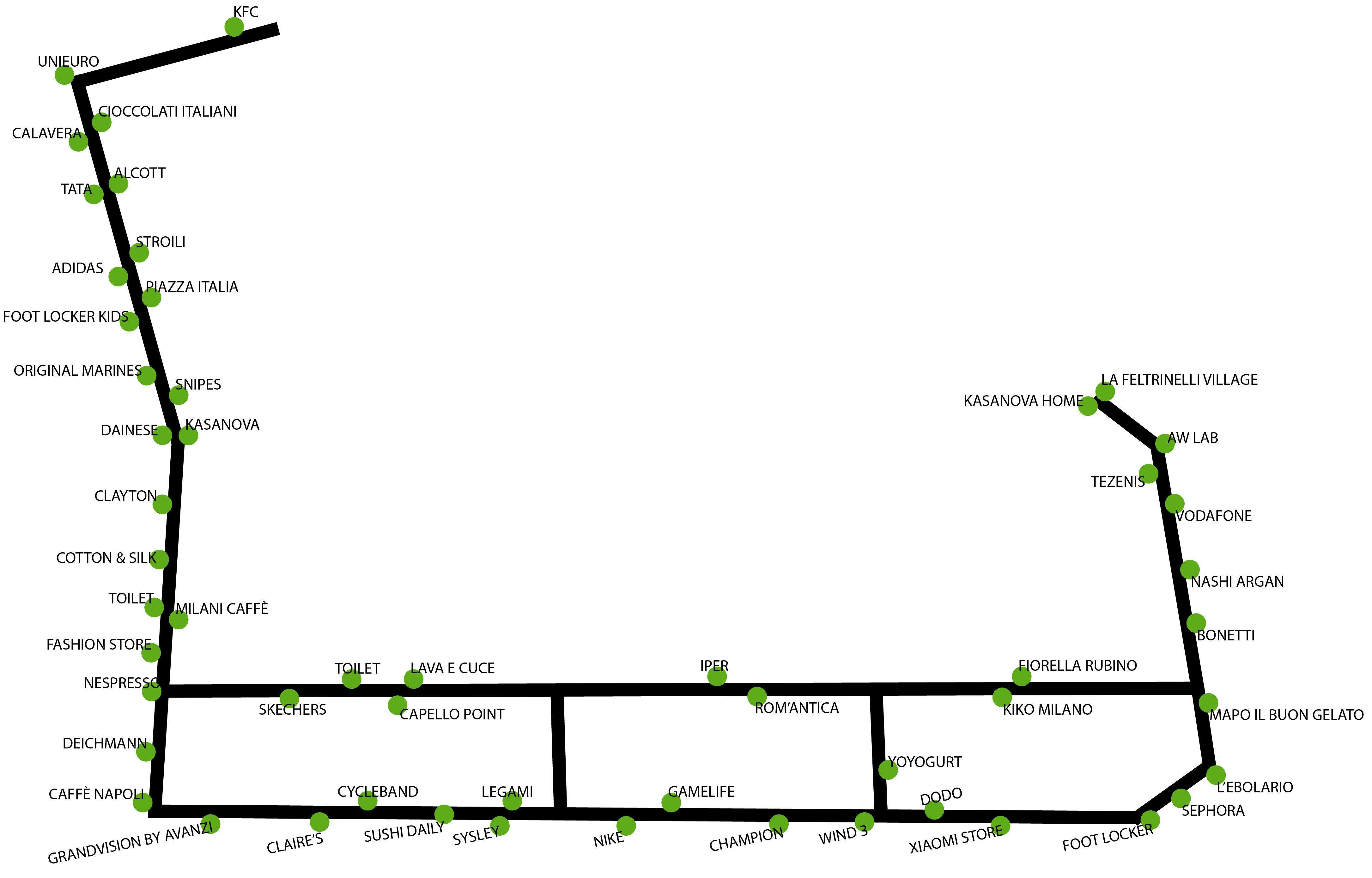}
  \caption{Graph-style image of Orio Center mall after Pre-processing phase.}
  \label{fig:bologna_path}
\end{figure}

\begin{itemize}
    \item \textit{Pre-processing}: the original images from Experiment I were manually replaced with a simplified graphical representation consisting solely of corridors and POIs as in the Figure \ref{fig:bologna_path}. Corridors are depicted as black lines, while POIs are represented by green dots labeled with their corresponding ID or name. This was carried out to increase the readability of the image by the model.
    The use of numbers, characters, or words to identify POIs in the new graph-style map reflects the conventions of the original maps. Legends using alphanumeric references were preserved as such, while named locations remained textual. Icons were converted into verbal descriptions
    \item \textit{System prompt}: \textit{SP\#1}, same as the previous Experiment I).
\end{itemize}

\begin{table*}[!ht]
\centering
\caption{Results with graph-style images and System prompt SP\#1 with ChatGPT-4o}
\label{second_test_table}
\begin{tabular}{lcccccc}
\toprule
\textbf{Query} 
& \multicolumn{2}{c}{\textbf{BERGAMO}} 
& \multicolumn{2}{c}{\textbf{BOLOGNA}} 
& \multicolumn{2}{c}{\textbf{ORIO}} \\
\cmidrule(lr){2-3} \cmidrule(lr){4-5} \cmidrule(lr){6-7}
& \textbf{Incorrect} & \textbf{Correct} 
& \textbf{Incorrect} & \textbf{Correct} 
& \textbf{Incorrect} & \textbf{Correct} \\
\midrule
Query-1 (S)     & 0 (0.00\%) & 7 (100.00\%)  & 4 (40.00\%) & 6 (60.00\%) & 4 (57.14\%) & 3 (42.86\%)\\
Query-2 (M)     & 1 (16.67\%) & 5 (83.33\%) & 2 (14.29\%) & 12 (85.71\%) & 7 (43.75\%) & 9 (56.25\%)\\
Query-3 (L)     & 5 (33.33\%) & 10 (66.67\%) & 4 (23.53\%) & 13 (76.47\%)  & 9 (40.91\%) & 13 (59.09\%)\\
Query-4 (Spec)  & 1 (20.00\%) & 4 (80.00\%) & 1 (12.50\%) & 7 (87.50\%)  & 1 (10.00\%) & 9 (90.00\%)\\
Query-5 (Spec)  & 6 (27.27\%) & 16 (72.73\%) & 1 (5.56\%) & 17 (94.44\%) & 6 (35.29\%) & 11 (64.71\%)\\
\midrule
\textbf{Total}  
& 13 \textbf{(23.64\%)} & 42 \textbf{(76.36\%)} 
& 12 \textbf{(17.91\%)} & 55 \textbf{(82.09\%)} 
& 27 \textbf{(37.50\%)} & 45 \textbf{(62.50\%)}\\
\bottomrule
\end{tabular}
\end{table*}

In Table \ref{second_test_table}, a total of 194 output indications were evaluated, with 52 incorrect (26.80\%), and 142 correct (73.20\%). The highest score of correct indications set is for Bologna Airport map (82.09\%) and the lowest result is for Orio-Center map (62.50\%).

There is a clear improvement, from which it can be deduced that stylized, graph-like maps perform better, thanks to the more essential (and relevant) information included. The major improvement observed in these tests is that the logical path through the corridors is followed correctly. All POIs along the route are correctly identified and presented. The most frequent errors still concern orientation: the model often fails to identify the initial direction to take, or whether a POI is located to the right or left. This could still be a problem of map representation, or it might be an intrinsic logical limitation of the LLM's internal reasoning, preventing it from correctly interpreting directionality.

\subsection{Experiment III - Model Upgrade}
\label{sec:test3}
This pipeline was initially tested using the same language model employed in the previous two experiments. However, this model did not provide strong evidence of a significant improvement over Experiment II. Therefore, we opted for the most powerful model of OpenAI, known as ChatGPT-o3.
\begin{itemize}
    \item \textit{Pre-processing}: Same images as Experiment II, with the graph representation based on corridors and POI.
    \item \textit{System prompt}: defined as '\textit{SP\#2}'. The difference between this and \textit{SP\#1} is only the proper and clear description of the map given by input. 
\end{itemize}

\begin{table*}[!ht]
\centering
\caption{Results with graph-style images and system prompt SP\#2 using ChatGPT-o3 (with reasoning)}
\label{final_test_table}
\begin{tabular}{lcccccc}
\toprule
\textbf{Query} 
& \multicolumn{2}{c}{\textbf{BERGAMO}} 
& \multicolumn{2}{c}{\textbf{BOLOGNA}} 
& \multicolumn{2}{c}{\textbf{ORIO}} \\
\cmidrule(lr){2-3} \cmidrule(lr){4-5} \cmidrule(lr){6-7}
& \textbf{Incorrect} & \textbf{Correct} 
& \textbf{Incorrect} & \textbf{Correct}  
& \textbf{Incorrect} & \textbf{Correct} \\
\midrule
Query-1 (S)     & 1 (10.00\%) & 9 (90.00\%)    & 2 (18.18\%)   & 9 (81.82\%)      & 1 (11.11\%) & 8 (88.89\%) \\
Query-2 (M)     & 1 (5.88\%) & 16 (94.12\%)    & 2 (6.67\%)    & 2 (93.33\%)      & 2 (8.70\%) & 21 (91.30\%) \\
Query-3 (L)     & 1 (2.86\%) & 34 (97.14\%)    & 8 (28.57\%)   & 20 (71.43\%)     & 7 (18.42\%) & 31 (81.58\%)\\
Query-4 (Spec)  & 2 (28.57\%) & 5 (71.43\%)    & 2 (13.33\%)   & 13 (86.67\%)     & 1 (5.26\%) & 18 (94.74\%)\\
Query-5 (Spec)  & 4 (14.81\%) & 23 (85.19\%)   & 5 (15.15\%)   & 28 (84.85\%)     & 5 (19.23\%) & 21 (80.77\%)\\
\midrule
\textbf{Total}  
& 9 \textbf{(9.38\%)} & 87 \textbf{(90.63\%)} 
& 19 \textbf{(16.24\%)} & 98 \textbf{(83.76\%)} 
& 16 \textbf{(13.91\%)} & 99 \textbf{(86.09\%)}\\
\bottomrule
\end{tabular}
\end{table*}

In Table \ref{final_test_table}, a total of 328 output indications were evaluated, with 44 incorrect (13.41\%), and 284 correct (86.59\%). The highest average score of correct indications set is for Bergamo Airport map (90.63\%) and the lowest result is with Bologna Airport map (83.76\%). With a peak of performance up to 97.14\% in Ber-3 query.

A relevant increase in the number of correct navigational indications was observed, particularly with a higher percentage of correctly identified left and right turns. The model demonstrated an improved sense of orientation throughout the path, accurately signaling directions all along the path.

Our findings demonstrate that the model achieves high levels of accuracy and reasoning performance, significantly outperforming the first two experiments.

\section{Discussion and conclusion}
 
\begin{table}
\centering
\small 
\setlength{\tabcolsep}{4pt} 
\renewcommand{\arraystretch}{0.9} 
\caption{Overall average results over tests}
\label{tab:total}
\begin{tabular}{lccc}
\toprule
\textbf{Experiment} & \textbf{Incorrect}  & \textbf{Correct} \\
\midrule
Experiment I & 49.46\%  & 50.54\% \\
Experiment II & 26.80\% & 73.20\% \\
Experiment III & 13.41\% & 86.59\% \\
\bottomrule
\end{tabular}
\end{table}




This paper evaluates the performance of ChatGPT in providing human-like indications to users in indoor environments represented through visual maps. 
We considered three different real-world map layouts with different characteristics, five types of engineered user prompts of different complexity levels, and three experimental settings, considering two different LLMs and pipeline strategies. In the three experiments, a total of 708 output indications were evaluated. The overall performance across all different experiments is summarized in Table~\ref{tab:total}. The results indicate that, with ChatGPT4-o3 and a stylized representation of the indoor maps (Experiment III), we can achieve 86\% correct indications, which is a highly promising result. 

Given the challenges of indoor navigation, this research has several key implications for AI-driven navigation, human-computer interaction, and assistive technologies. Users often ask questions such as ``What can I find on this floor?" or ``Where is the nearest restroom?" highlighting the need for adaptable navigation solutions that can dynamically respond to different environments and user preferences. A detailed and accurate description of the surrounding environment can enhance spatial awareness and improve the overall navigation experience. The use of LLMs can substantially reduce the costs of an indoor navigation system, as it relies solely on software. Furthermore, the LLM requires only a pre-processed, graph-like map of the target building, which speeds-up the deployment in diverse environments, even in the presence of layout updates (e.g., due to work-in-progress, or expansion of the building). In our envisioned system, users can use their mobiles to interact with an LLM-based assistant, which can guide them to their desired destination efficiently, especially if the LLM can also leverage wi-fi fingerprints and provide instructions considering the current position of the user. The use of wi-fi can also help to correct the errors introduced by the LLMs (e.g., confusion between left and right), thereby improving the quality of the indications. 

A critical limitation of the best solution identified is the average response time—approximately five minutes per query—posing a substantial barrier to real-time usage. Future research directions include exploring more optimized inference strategies or lightweight alternatives to reduce latency without compromising the quality of reasoning. Another challenge to address is the automatic translation of the user query into an engineered prompt, which can, in principle, be achieved again through the mediation of the LLM. 

Besides addressing the current limitations, our future work will be oriented to evolve the idea to support users with different accessibility problems, also through the integration with localization systems. 

Future iterations could integrate real-time obstacle detection via smartphone sensors or connected assistive devices (e.g., smart canes) to supplement LLM-generated instructions. For individuals with motor disabilities, the system can prioritize accessible routes 
and adapt instructions based on mobility aids. 

Additionally, voice-controlled navigation and hands-free querying 
can reduce physical interaction demands. 

In conclusion, our paper is the first step to make indoor navigation not only similar to outdoor navigation systems in terms of user experience, but also more inclusive towards diverse accessibility needs while reducing hardware dependencies.


\bibliographystyle{IEEEtran} 

\bibliography{bibliography}

\begin{thebibliography}{10}
\providecommand{\url}[1]{#1}
\csname url@samestyle\endcsname
\providecommand{\newblock}{\relax}
\providecommand{\bibinfo}[2]{#2}
\providecommand{\BIBentrySTDinterwordspacing}{\spaceskip=0pt\relax}
\providecommand{\BIBentryALTinterwordstretchfactor}{4}
\providecommand{\BIBentryALTinterwordspacing}{\spaceskip=\fontdimen2\font plus
\BIBentryALTinterwordstretchfactor\fontdimen3\font minus \fontdimen4\font\relax}
\providecommand{\BIBforeignlanguage}[2]{{%
\expandafter\ifx\csname l@#1\endcsname\relax
\typeout{** WARNING: IEEEtran.bst: No hyphenation pattern has been}%
\typeout{** loaded for the language `#1'. Using the pattern for}%
\typeout{** the default language instead.}%
\else
\language=\csname l@#1\endcsname
\fi
#2}}
\providecommand{\BIBdecl}{\relax}
\BIBdecl

\bibitem{torres2021towards}
J.~Torres-Sospedra, I.~Silva, L.~Klus, D.~Quezada-Gaibor, A.~Crivello, P.~Barsocchi, C.~Pend{\~a}o, E.~S. Lohan, J.~Nurmi, and A.~Moreira, ``Towards ubiquitous indoor positioning: Comparing systems across heterogeneous datasets,'' in \emph{2021 International Conference on Indoor Positioning and Indoor Navigation (IPIN)}.\hskip 1em plus 0.5em minus 0.4em\relax IEEE, 2021, pp. 1--8.

\bibitem{furfari2019next}
F.~Furfari, A.~Crivello, P.~Barsocchi, F.~Palumbo, and F.~Potort{\`\i}, ``What is next for indoor localisation? taxonomy, protocols, and patterns for advanced location based services,'' in \emph{2019 International Conference on Indoor Positioning and Indoor Navigation (IPIN)}.\hskip 1em plus 0.5em minus 0.4em\relax IEEE, 2019, pp. 1--8.

\bibitem{shao2024moc}
W.~Shao, H.~Luo, F.~Zhao, Y.~Hong, Y.~Li, C.~Zhang, B.~Sun, and A.~Crivello, ``Moc: Wi-fi ftm with motion observation chain for pervasive indoor positioning,'' \emph{IEEE Transactions on Industrial Informatics}, 2024.

\bibitem{buzzi2024chatbot}
M.~Buzzi, E.~Grassini, and B.~Leporini, ``A chatbot-based assistive technology to get information on a clinical environment,'' in \emph{Proceedings of the 17th International Conference on PErvasive Technologies Related to Assistive Environments}, 2024, pp. 504--509.

\bibitem{tsai2023multimodal}
Y.-H.~H. Tsai, V.~Dhar, J.~Li, B.~Zhang, and J.~Zhang, ``Multimodal large language model for visual navigation,'' \emph{arXiv preprint arXiv:2310.08669}, 2023.

\bibitem{fang2024travellmplannewpublic}
\BIBentryALTinterwordspacing
B.~Fang, Z.~Yang, S.~Wang, and X.~Di, ``Travellm: Could you plan my new public transit route in face of a network disruption?'' 2024. [Online]. Available: \url{https://arxiv.org/abs/2407.14926}
\BIBentrySTDinterwordspacing

\bibitem{10.1145/3688828.3699636}
\BIBentryALTinterwordspacing
H.~Zhang, N.~J. Falletta, J.~Xie, R.~Yu, S.~Lee, S.~M. Billah, and J.~M. Carroll, ``Enhancing the travel experience for people with visual impairments through multimodal interaction: Navigpt, a real-time ai-driven mobile navigation system,'' in \emph{Companion Proceedings of the 2025 ACM International Conference on Supporting Group Work}, ser. GROUP '25.\hskip 1em plus 0.5em minus 0.4em\relax New York, NY, USA: Association for Computing Machinery, 2025, p. 29–35. [Online]. Available: \url{https://doi.org/10.1145/3688828.3699636}
\BIBentrySTDinterwordspacing

\bibitem{app14209343}
\BIBentryALTinterwordspacing
K.~Lipka, D.~Gotlib, and K.~Choromański, ``The use of language models to support the development of cartographic descriptions of a building’s interior,'' \emph{Applied Sciences}, vol.~14, no.~20, 2024. [Online]. Available: \url{https://www.mdpi.com/2076-3417/14/20/9343}
\BIBentrySTDinterwordspacing

\bibitem{anagnostopoulos2025ordip}
G.~G. Anagnostopoulos, P.~Barsocchi, A.~Crivello, C.~Pend{\~a}o, I.~Silva, and J.~Torres-Sospedra, ``Ordip: Principle, practice and guidelines for open research data in indoor positioning,'' \emph{Internet of Things}, p. 101485, 2025.

\bibitem{coffrini2025methodllmenabledindoornavigation}
\BIBentryALTinterwordspacing
A.~Coffrini, M.~A. Zadenoori, P.~Barsocchi, F.~Furfari, A.~Crivello, and A.~Ferrari, ``Toward a method for llm-enabled indoor navigation,'' 2025. [Online]. Available: \url{https://arxiv.org/abs/2503.11702}
\BIBentrySTDinterwordspacing

\end{thebibliography}

\end{document}